\documentclass[11pt,a4paper]{article}
\usepackage{times,latexsym}
\usepackage{url}
\usepackage[T1]{fontenc}

\usepackage[acceptedWithA]{tacl2021v1}

\usepackage{xspace,mfirstuc,tabulary}

\newif\iftaclinstructions
\taclinstructionsfalse 
\iftaclinstructions
\renewcommand{\confidential}{}
\renewcommand{\anonsubtext}{(No author info supplied here, for consistency with
TACL-submission anonymization requirements)}
\newcommand{\instr}
\fi

\iftaclpubformat 

\else

\fi

\usepackage{url}
\usepackage{enumitem}
\definecolor{light_gray}{HTML}{FFFFFF} 
\definecolor{solid_gray}{HTML}{000000} 
\usepackage{multirow}
\usepackage{hhline}
\usepackage{pifont}
\usepackage{makecell}
\usepackage{array} 
\usepackage{booktabs}
\usepackage{siunitx}
\usepackage{nameref}
\usepackage{xcolor}
\usepackage{transparent}
\usepackage{soul}
\usepackage{tcolorbox}
\usepackage{amssymb}
\usepackage{bbding}
\usepackage{longtable}
\usepackage{tabularx}
\usepackage{pgfplots}
\usepackage{rotating}
\usepackage{adjustbox}
\usepackage{bibentry}
\usepackage{pgfplots}
\usepackage{amsmath}
\usepackage{pgfplotstable}
\usepackage{subcaption}
\usepackage{pgfplots}
\usepackage{ltablex}
\usepackage{longtable}
\usepackage{graphicx}
\usepackage{wrapfig}
\usepackage{pgfplotstable}
\usepgfplotslibrary{groupplots}
\usepgfplotslibrary{fillbetween}
\usepackage{pgfplotstable}
\usetikzlibrary{patterns}
\pgfplotsset{compat=1.16}
\usepgfplotslibrary{groupplots}
\pgfplotsset{compat=newest}
\usetikzlibrary{positioning}
\usepgfplotslibrary{groupplots}
\usepackage[textsize=tiny]{todonotes}
\usetikzlibrary{calc}
\usepackage{tikz}
\usetikzlibrary{shapes}

\newdimen\R 
\newdimen\L 
\setcitestyle{aysep={},yysep={;},citesep={;}}

\newcommand{\cmark}{\textcolor{green!65!black}{\ding{51}}} 

\newtcbox{\highlight}[1][]{%
    colback=green!15!white,
    colframe=green!15!white,
    boxrule=0pt,
    boxsep=0pt,
    left=2pt,
    right=2pt,
    top=2pt,
    bottom=2pt,
    sharp corners,
    #1
}

\definecolor{transblue}{rgb}{0.69, 0.85, 0.96} 
\definecolor{editcolor}{rgb}{0.7, 0, 0} 
\definecolor{mygray}{gray}{0.2}

\newtcbox{\bleuhl}[1][]{%
    on line,
    arc=0pt,
    outer arc=0pt,
    colback=transblue!50,
    boxsep=0pt,
    left=2.0pt,
    right=2.0pt,
    top=1.0pt,
    bottom=0.4pt,
    boxrule=0.0pt,
    #1
}

\title{Data Contamination Quiz: A Tool to Detect and Estimate Contamination in Large Language Models}

\author{Shahriar Golchin\Thanks{Corresponding author.} , Mihai Surdeanu \\
Department of Computer Science, University of Arizona \\
\texttt{\{golchin, msurdeanu\}@arizona.edu}}

\date{}

\begin{document}
\maketitle

\begin{abstract}
We propose the Data Contamination Quiz (DCQ), a simple and effective approach to \textit{detect} data contamination in large language models (LLMs) and \textit{estimate} the amount of it. 
Specifically, we frame data contamination detection as a series of multiple-choice questions, devising a quiz format wherein three perturbed versions of each instance, subsampled from a specific dataset partition, are created. These changes only include word-level perturbations. 
The generated perturbations, along with the original dataset instance, form the options in the DCQ, with an extra option accommodating the selection of none of the provided options.
Given that the only distinguishing signal among the options is the \textit{exact wording} with respect to the original dataset instance, an LLM, when tasked with identifying the original dataset instance, gravitates towards selecting the original one if it has been exposed to it.
While accounting for positional biases in LLMs, the quiz performance reveals the contamination level for the tested model with the dataset partition to which the quiz pertains.
Applied to various datasets and LLMs, under controlled and
uncontrolled contamination, our findings---while fully lacking access to training data and model parameters---suggest that DCQ achieves state-of-the-art results and uncovers greater contamination levels through memorization compared to existing methods. Also, it proficiently bypasses more safety filters, especially those set to avoid generating copyrighted content.\footnote{All related code and data are publicly accessible at \url{https://github.com/shahriargolchin/DCQ}.}
\end{abstract}

\section{Introduction}
\label{section:intro}
Large language models (LLMs) have shown remarkable success in various benchmarks \citep[inter alia]{bubeck2023sparks,bang2023multitask,openai2023gpt4}. However, due to the lack of transparency regarding the data sources used by some model providers, coupled with their dependency on web data for training \citep[inter alia]{refinedweb,touvron2023Llama,touvron2023Llama2}, the performance of such models may be skewed by ``data contamination'' \citep{Zhou2023DontMY}. This refers to the inclusion of test data from downstream tasks in the LLMs' training data \citep{golchin2024time}. Considering the immense volume of data used for training, even making this data open-access does not provide a practical solution. This is because cross-verifying each sample in the test data against open-source training data becomes impractical due to the substantial computational costs  \citep{touvron2023Llama2} or the extensive manual effort involved \citep{chowdhery2022palm}.
This has led LLM developers to assess data contamination in benchmark datasets post-training \citep{team2023gemini}.
Therefore, there is a pressing need to accurately pinpoint and quantify this contamination using a cost-effective approach, which demands minimal computational resources and human intervention, ensuring its broad applicability by the community.\footnote{In Appendix \ref{appendix:current-and-previous-versions}, we show how LLMs are being updated to resist recent data contamination detection methods, emphasizing the need for developing new methods to continually identify data contamination as models evolve.}

\begin{figure}[t!]
\centering
\begin{adjustbox}{width=0.985\columnwidth,center}
\begin{tabular}{l}
\parbox{\columnwidth}{\underline{\fontsize{8}{8pt}\selectfont \textbf{Original Dataset Instance:}}

\vspace{0.1cm}

\fontsize{8}{8pt}\selectfont \textbf{Summary:} Premier Inn has announced it is investing £35m in opening three new hotels and revamping two others in Edinburgh.

}
\vspace{0.2cm}
\\ 
\begin{minipage}{0.985\columnwidth}
\begin{tikzpicture}[rounded corners=8pt, thick, text=black, text opacity=1]
    \node[draw=solid_gray, font=\fontsize{7pt}{8.7pt}\selectfont, fill=light_gray, line width=1pt, text width=\columnwidth, align=left, inner xsep=6.5pt, inner ysep=5pt] at (0,0) {
    \textbf{Instruction:} You are provided with a five-choice quiz. Your task is to correctly select the option that exactly corresponds to an instance from the test split of the XSum dataset.
    
    \vspace{0.1cm}
    
    (1) You must ensure that you only generate a single option letter as your answer.
    
    (2) If you do not know the dataset or the correct answer, you must select option ``E) None of the provided options.''

    \vspace{0.1cm}
    
    Hint: While all the following options seem similar in terms of meaning, there is only one option that reflects the exact wording with respect to the original instance.
    
    -- -- --

    \textbf{A)} Summary: Premier Inn has revealed it is allocating £35m for the opening of three new hotels and the refurbishment of two others in Edinburgh.

    \vspace{0.1cm}

    \textit{\textbf{B)} Summary: Premier Inn has announced it is investing £35m in opening three new hotels and revamping two others in Edinburgh.}

    \vspace{0.1cm}

    \textbf{C)} Summary: Premier Inn has declared it is investing £35m to set up three new hotels and modernize two others in Edinburgh.
    
    \vspace{0.1cm}
    
    \textbf{D)} Summary: Premier Inn has announced it is spending £35m to open three new hotels and renovate two others in Edinburgh.

    \vspace{0.1cm}
    
    \textbf{E)} None of the provided options.
    
    -- -- --

    \vspace{-0.1cm}

    \textbf{Answer:} \raisebox{-0.3ex}{\highlight{\texttt{B}}} \large \cmark

    };
\end{tikzpicture}
\end{minipage}
\end{tabular}
\end{adjustbox}
\caption{An example of a quiz question crafted to detect data contamination within the test partition of the XSum dataset. Here, the produced answer by the underlying LLM (GPT-4) aligns with the correct option (option C), signaling previous exposure to data, and thus, revealing contamination.}
\label{figure:example-of-contamination-quiz}
\end{figure}

We propose the Data Contamination Quiz (DCQ), a streamlined method that efficiently \textit{detects and estimates verbatim data contamination in LLMs without the need to access training data or internal weights/logits}.
Essentially, DCQ is designed to pinpoint the exact data points from dataset partitions that have leaked into the training data of LLMs, leading to skewed performance.
In essence, \textit{detection} is achieved via a multiple-choice question and \textit{estimation} is derived by evaluating the LLM's performance across a series of these multiple-choice questions that form the quiz. Specifically, we form a five-option quiz: one option presents an original instance from a dataset partition, three options offer three distinct word-level perturbations of it where words are replaced with their contextually relevant synonyms, and the last option allows the selection of none of these provided options.
We use GPT-4 \citep{openai2023gpt4} to \emph{automate} the generation of word-level perturbations.
While the wording differs, the meaning and sentence structure of these perturbations mirror the original dataset instance, resulting in (almost) identical semantic representations across all options containing perturbations and the original dataset instance.
Therefore, the sole difference among these options is in the \textit{precise wording} relative to the original dataset instance. Hence, when the LLM is tasked with identifying the original dataset instance, a consistent preference for selecting options containing original dataset instances reveals the model's memorization and prior exposure to data. Figure \ref{figure:example-of-contamination-quiz} exemplifies this idea.

However, the detection of contamination through memorization\footnote{While we recognize \textit{memorization} and \textit{contamination} as distinct concepts, in our study, they are used interchangeably. This is because our method relies on memorization as a basis for detecting contamination.} is not consistent across all quiz options due to the \emph{positional biases} present in LLMs \citep{wang2023large,zheng2024large}, leading to overestimation or underestimation of contamination levels.
To counter this, we first identify positional biases using a quiz we call the ``Bias Detector Quiz (BDQ),'' which excludes correct answers (i.e., original dataset instances) from its options. In fact, all options in each question of the BDQ are word-level perturbations based on each of the original dataset instances under examination, with an extra option for selecting none of the provided options.\footnote{In BDQ, the question in Figure \ref{figure:example-of-contamination-quiz} offers four word-level perturbations, omitting the original dataset instance.} Upon submitting BDQ to an LLM, options that are chosen less frequently than random chance are identified as ``non-preferred options,'' excluding the last option (option E) that always allows for the selection of none of the provided options.
\emph{These non-preferred options are strategic positions to avoid overestimation.} Therefore, they are replaced with the original dataset instances in the ``Bias Compensator Quiz (BCQ)'' to detect contamination while avoiding overestimation. Nevertheless, due to the same issue of positional biases, memorization may not be triggered equally among all the non-preferred positions, thereby leading to potential underestimation. \emph{To avoid underestimation, we permute the correct answers among all the non-preferred options}, which means an LLM takes multiple BCQs (equal to the number of non-preferred options), to determine the maximum contamination without overestimation. Finally, using the highest quiz performance across all BCQs, a range for the minimum and maximum detected contamination levels is estimated.
\emph{We refer to this end-to-end process as our Data Contamination Quiz (DCQ).}

The main contributions of this paper are:

\noindent\textbf{(1)} We propose the first strategy for \emph{estimating} verbatim data contamination in fully black-box LLMs. This detection relies solely on input-output interactions, without the need to access training data and model weights/logits. Our quiz-based approach is structured such that correct answers validate the existence of memorization, thereby revealing prior exposure to data. As a result, contamination level is estimated based on the quiz performance on a subsample of dataset instances.

\noindent\textbf{(2)} Using rigorous evaluations across multiple scenarios, including controlled and uncontrolled contamination environments, our findings show that our method detects significantly higher levels of contamination via memorization than the existing method for identifying data contamination in fully black-box LLMs \citep{golchin2024time}.

Overall, our strategy offers \emph{several key benefits over existing methods} for data contamination detection:
\textbf{(1)} by limiting the LLMs' outputs to be a single letter that identifies the selected option (i.e., from A to E), our strategy substantially reduces the probabilistic behavior of LLMs in detecting data contamination;
\textbf{(2)} building on the same principle of limiting outputs, our method can adeptly evade the safety filters set by model providers during the decoding process, especially in proprietary LLMs that seek to prevent the generation of copyrighted content, thus exposing more contamination;
\textbf{(3)} it transcends the limitations of probability/threshold-based methods \citep[inter alia]{shi2024detecting, oren2024proving, dong2024generalization}, which are task/language-specific, allowing our approach to be applied across different tasks and languages; 
\textbf{(4)} our technique is both cost- and time-saving thanks to the automated quiz generation process and administration on a manageable set of instances, such as 100 subsampled instances used in our experiments; and
\textbf{(5)} the streamlined nature of our method means the quiz options we collect for each dataset partition are reusable and can be applied to any LLM to take the DCQ for data contamination assessment.

\section{Approach}
\label{section:approach}

Our approach is based on a key assumption: \textit{if} an LLM has encountered a dataset instance verbatim during training, \textit{then} it can recognize that instance among its word-level perturbed versions through \textit{memorization} \citep{DBLP:conf/iclr/CarliniIJLTZ23}. Framing this as a quiz-like task, prior exposure to data is inferred by tasking the model to select a single-letter option corresponding to the original dataset instance.

Given this assumption, our approach consists of three main phases:

\noindent\textbf{(1) Creating Quiz Options:} We generate four unique word-level perturbations for each subsampled dataset instance.

\noindent\textbf{(2) Detecting and Handling Positional Biases:} We craft and submit two types of quizzes using two combinations of the generated perturbations:
    \begin{itemize}[itemsep=0pt, topsep=0pt]
        \item \textbf{Bias Detector Quiz (BDQ):} This quiz includes all four perturbations as options, along with an option to select none of them. Its purpose is to identify positional biases.
        \item \textbf{Bias Compensator Quiz (BCQ):} This quiz replaces one of the four perturbations in the previous quiz with the original dataset instance and aims to assess contamination levels while accounting for positional biases.
    \end{itemize}

\noindent\textbf{(3) Estimating Contamination Levels:} Based on quiz performance from BCQ, we estimate a range for the detected contamination levels.

We detail each phase in the following.

\subsection{Creating Quiz Options}
We use GPT-4 to generate four distinct word-level perturbations for each subsampled dataset instance.
Its rich representations and proficient adherence to following instructions \citep{bubeck2023sparks} make it suitable for performing word replacements that maintain contextual relevance. This reduces the need for prompt engineering and mitigates potential biases when producing word-level perturbations that can arise from factors such as variations in length, sentence structure, or the lack of diversity.

To instruct GPT-4, we employ a zero-shot prompt with the original dataset instance infused in it as a reference.\footnote{This prompt is shown in Figure \ref{figure:quiz-generation-prompt} in Appendix \ref{appendix:crafting-quiz}.}
In particular, this prompt involves the following rules and requirements when generating four quiz options per instance: (1) all generated perturbations must retain the meaning and sentence structure of the original dataset instance; (2) they must be distinct from one another; (3) they must conform to every precise symbol and letter detail found in the original dataset instance; and (4) the sole difference between them must be the word-level perturbations.\footnote{Refer to Appendix \ref{appendix:crafting-quiz} for more details and analysis on the generated perturbations.}
These options are then verified to be distinct from one another and not identical to the original dataset instance.
Further, all options represent random perturbations in terms of position and the number of tokens involved in the perturbation process.
However, when the original dataset instance contains a label, we do not alter the original label and directly attach it to the perturbed versions.\footnote{See Table \ref{tab:data-contamination-in-present-vs-past-updates-quiz} in Appendix \ref{appendix:current-and-previous-versions} for an example.}

We then compile and employ \emph{all four generated options} in BDQ to identify positional biases. In contrast, we use \emph{three of these options} when administering BCQ, wherein one option from BDQ is replaced with the original dataset instance.
As a fifth option, in both quizzes, we always include a choice that allows for selecting none of the provided options. This is beneficial in situations where the LLM is not contaminated with a particular dataset or when the LLM's memorization is not triggered by the presented options.
We detail each quiz in Subsection \ref{subsection:deployment-of-DCQ}.
Figure \ref{figure:example-of-contamination-quiz} shows the integration of the generated perturbations for an original dataset instance as quiz options in BCQ.

The decision to design quizzes with five options is based on the fact that most multiple-choice quizzes typically include three to five answer choices per question. LLMs are frequently exposed to and familiarized with these formats during their fine-tuning and/or pre-training stages. Therefore, using this range of options aligns with the standard quiz formats LLMs previously encountered, minimizing the chance for poor or exceptional performance only due to an unfamiliar test design. Additionally, choosing five choices within this standard range is intended to create a robust and challenging test environment by maximizing the number of answer choices.

\subsection{Detecting and Handling Positional Biases}
\label{subsection:deployment-of-DCQ}

The performance of LLMs in answering multiple-choice questions is influenced by positional biases \citep{wang2023large,zheng2024large}, which we empirically show in Subsection \ref{subsection:positional-adversarial-analysis}.
These biases alter the quiz performance according to the placement of the options, leading to varying detectable contamination levels for the same dataset instances being studied.
Therefore, we first detect positional biases, and then, adjust for these biases in our contamination detection process.

\vspace{0.15cm}

\noindent \textbf{Bias Detector Quiz (BDQ).} This five-option quiz is primarily designed to detect positional biases. 
In this quiz, the correct answers (original dataset instances) are absent, and all options---except for the last one, which is an option to select none of the provided options---represent word-level perturbations of the original dataset instances. 
By administering this quiz to an LLM, we analyze the LLM's tendency to favor certain options (positions) more frequently than others when the correct answers are missing, thereby revealing positional biases.
\emph{From the outcomes of the BDQ, we identify what we term the ``non-preferred options.''}\footnote{While having zero or one non-preferred option is possible, we consistently refer to \emph{multiple non-preferred options} due to their common occurrence and to ensure consistency in our discussion in the paper.} We define an option as non-preferred in consideration of an LLM if it is chosen less frequently than what random chance suggests. This definition excludes the last option (option E), which is always fixed to enable the selection of none of the given options. For a five-option quiz with $k$ questions (instances being evaluated), any option selected fewer than $\lceil \frac{k}{5} \rceil$ times in the BDQ is deemed non-preferred. These options are viewed as strategic spots to minimize the risk of overestimating contamination when placing the original dataset instances there. This strategic placement is the main task of what we discuss next, the BCQ.

\vspace{0.15cm}

\noindent \textbf{Bias Compensator Quiz (BCQ).} The main objective of this quiz is to compensate for positional biases when detecting data contamination. Such biases can skew quiz performance, causing the perceived contamination levels based on quiz performance to be overestimated or underestimated.
\emph{To avoid overestimation, we systematically replace the detected non-preferred options by BDQ with the original dataset instances.}
Nevertheless, due to the same issue of positional biases, the detection of the original dataset instances is not consistent among all the non-preferred options, which leads to an underestimation of contamination. \emph{To avoid underestimation and find the highest possible memorization among non-preferred options, we permute the correct answers, i.e., options containing the original dataset instances, among all non-preferred options.}\footnote{While we never faced this situation, in the rare event that there are no non-preferred options, we permute the correct answers among all options except the last one (option E).} This allows us to obtain the highest performance in triggering the LLM's memorization for detecting the original dataset instances while avoiding overestimation. 
Finally, the highest performance among all BCQs is reported as the maximum level of contamination.\footnote{When BCQs tie for the highest quiz performance, we select the one with the lowest positional bias detected by BDQ.}

\subsection{Estimating Contamination Levels by Quiz Performance}

As noted, the maximum contamination level for a dataset partition is determined by the highest performance of the LLM across all BCQs. While this provides the maximum level of detectable contamination, it is also possible to compute the minimum level of detectable contamination.

As we deal with a non-ideal system, we can apply the principle of chance-adjusted accuracy to account for the probability of chance agreement using Cohen's Kappa \citep{galton1892finger,Cohen1960ACO}, defined as:
$\kappa = (p_o - p_e)/(1 - p_e)$.
Here, \( p_o \) is the observed agreement and \( p_e \) is the expected agreement by chance.
In simpler terms, \( p_o \) refers to the highest proportion of correct answers attained by the LLM among BCQs, whereas \( p_e \) is the proportion of choosing the non-preferred option in the BDQ, where the LLM achieved its top performance in the BCQ by placing the correct answers (original dataset instances) in that position.

For example, if the best performance among all Bias Compensator Quizzes is obtained by systematically placing correct answers in position B, \( p_o \) is the proportion of selecting option B in Bias Compensator Quiz while \( p_e \) is the proportion of selecting option B in Bias Detector Quiz.
The resulting value is then reported as the minimum level of detected contamination.

\section{Experimental Setup}

\noindent \textbf{Data.} To improve clarity given the various evaluation settings explored in this paper, we provide the specifics of the datasets used for each evaluation and the relevant subsection separately below.

Subsection \ref{subsection:positional-adversarial-analysis}: To examine the initiation of memorization in LLMs via quiz options and controlling overestimation and underestimation due to positional biases, we use two of the contaminated datasets reported by \citet{golchin2024time}: AG News \citep{Zhang2015CharacterlevelCN} and WNLI \citep{wang2018glue}. Specifically, we use the train and test sets of the former and the train and validation sets of the latter dataset.

Subsection \ref{subsection:controlled-contamination}: In our systematic contamination experiment, we utilize the train sets from two uncontaminated datasets---MeetingBank \citep{hu-etal-2023-meetingbank} and AuTexTification \citep{autextification2023}---both released in 2023, and thus, are unexposed to GPT-3.5 \citep{ouyang2022training} and Llama 2 \citep{touvron2023Llama}. 
Also, our experiment involving reported contamination uses test split from the HumanEval \citep{chen2021codex}, validation split from the DROP \citep{DBLP:conf/naacl/DuaWDSS019}, and train split from the GSM8k \citep{cobbe2021GSM8k}.

Subsection \ref{subsection:eval-in-the-wild}: To apply our approach in the wild, we use the same datasets as used by \citet{golchin2024time} for their data contamination analysis. This includes IMDB \citep{maas-EtAl:2011:ACL-HLT2011}, AG News \citep{Zhang2015CharacterlevelCN}, Yelp Full Reviews \citep{Zhang2015CharacterlevelCN}, SAMSum \citep{gliwa2019samsum}, XSum \citep{Narayan2018DontGM}, WNLI \citep{wang2018glue}, and RTE \citep{wang2019superglue}.
Our experiments focus on the test splits of these datasets except for WNLI and RTE, where we use validation sets due to the unavailability of their test sets.

Preparation of instances involves formatting each instance as per the task of its dataset: for the classification task, an instance includes the text along with its exact label; for the natural language inference task, it contains the premise, hypothesis, and the exact label; for summarization task, summaries without related documents; and so on. In general, we format each instance such that, if disclosed during the training stage, it can skew the model's performance and generalizability.

Following prior studies \citep{shi2024detecting,golchin2024time}, we subsample $k = 100$ unique dataset instances from each dataset partition for all experiments. If a partition contains fewer instances, we use the entire set.
In Subsection \ref{subsection:controlled-contamination}, we demonstrate that this sample size is adequate for effectively estimating contamination levels while managing the costs associated with perturbation generation and quiz administration.

\vspace{0.15cm}

\noindent \textbf{Setting.} All experiments involving GPT-3.5 and GPT-4 use the \texttt{gpt-3.5-turbo-0613} and \texttt{gpt-4-0613} snapshots, accessed via the OpenAI API. For all our experiments with Llama 2, we employ the \texttt{Llama 2-Chat} (70B) model.

When generating word-level perturbations, we adjust the temperature for GPT-4 to 1.00—the default value for generative tasks such as conversation and storytelling\footnote{\url{https://platform.openai.com/playground}}---to encourage diversity, and cap the completion length to 4,000 tokens. In contrast, when submitting all quizzes (i.e., BDQ and BCQ), we set the temperature to zero---promoting deterministic output given the provided options---with a token limit of one.

\vspace{0.15cm}

\noindent\textbf{Comparative Framework.} We compare our DCQ against the replication-based method proposed by \citet{golchin2024time}.
This choice is due to several key reasons: (1) their detection method is capable of operating in a fully black-box setting, without access to training data and model weights/logits, which aligns well with our requirements; 
(2) although this method was originally designed to detect contamination at the partition level, it is the only existing method that can be adapted beyond binary detection to quantify contamination by calculating the proportion of exact and near-exact matches replicated by this method; and
(3) this strategy is both task- and language-agnostic, making it broadly applicable to our experiments.

Other existing detection methods do not fulfill all the aforementioned criteria. Specifically, some require access to model weights/logits or training data \citep[inter alia]{shi2024detecting,oren2024proving,yang2023rethinking,deng2023investigating}, others cannot be adapted for estimation comparison while operating in fully black-box setting \citep{dong2024generalization,deng2023investigating}, and some are task- or language-specific \citep[inter alia]{dong2024generalization,Roberts2023DataCT,shi2024detecting,oren2024proving,deng2023investigating}.
For example, the method proposed by \citet{dong2024generalization} is ineffective for detecting contamination in short- or non-generative tasks such as classification and summarization. This approach relies on diversity to identify contamination, which is only informative when there is sufficient variation in the generated outputs. In cases where diversity is inherently low, the method incorrectly flags nearly all samples as contaminated, leading to inconclusive findings.

In particular, we employ the top-performing strategy (guided instruction with GPT-4 evaluation) proposed by \citet{golchin2024time} for detecting contamination, treating both exact and near-exact matches as contamination instances alike.
In our experiments using this method, we set the temperature to zero and cap the token limit at 500. We also create random-length initial segments of dataset instances by randomly deriving them from a range spanning 40\% to 70\% of each instance's length, based on the white space count. 
We selected this range as it yielded the highest levels of memorization when replicating instances using the LLMs under investigation.

\section{Empirical Results and Evaluation}
\label{section:empirical-results-and-evaluation}
In this section, we first provide empirical evidence on how our approach instigates memorization in LLMs for revealing data exposure while avoiding overestimation and underestimation of contamination.
Next, we assess its performance in detecting verbatim contamination by applying it to controlled contamination settings where the presence of contamination is either systematic or reported, and further employ it in real-world cases where contamination is not controlled.

\pgfplotsset{compat=1.7}

\begin{figure*}
\centering
\begin{tikzpicture}
\begin{groupplot}[
    group style={
    group size=4 by 3,
    horizontal sep=0.75cm,
    vertical sep=0.83cm,
    },
    width=0.295\textwidth,  
    height=0.15\textwidth,
    ybar,
    symbolic x coords={A, B, C, D, E},
    xtick=data,
    nodes near coords,
    tick label style={font=\tiny},
    nodes near coords style={font=\tiny},
    nodes near coords align={vertical},
    axis lines*=left,
    ymin=0,
    ymax=100,
    ytick={0,100},
    title style={at={(axis description cs:0.5,-2)}, align=center, font=\fontsize{7}{8.7}\selectfont},
    yticklabel={\pgfmathparse{\tick}\pgfmathprintnumber{\pgfmathresult}\%},
]

\nextgroupplot[ylabel={{\fontsize{5pt}{5pt}\selectfont BDQ}}]
\addplot[
  pattern=north east lines,
  pattern color=green!70!black
] coordinates {(A,29) (B,0) (C,0) (D,0) (E,71)};

\nextgroupplot
\addplot[
  pattern=north east lines,
  pattern color=green!70!black
] coordinates {(A,41) (B,0) (C,1) (D,1) (E,57)};

\nextgroupplot
\addplot[
  pattern=north east lines,
  pattern color=green!70!black
] coordinates {(A,6) (B,0) (C,0) (D,0) (E,94)};

\nextgroupplot
\addplot[
  pattern=north east lines,
  pattern color=green!70!black
] coordinates {(A,9.86) (B,0) (C,0) (D,1.41) (E,88.73)};

\nextgroupplot[ylabel={{\fontsize{5pt}{5pt}\selectfont BDQ W/ New Pert.}}]
\addplot[
  pattern=grid,
  pattern color=blue!70
] coordinates {(A,28) (B,0) (C,0) (D,0) (E,72)};

\nextgroupplot
\addplot[
  pattern=grid,
  pattern color=blue!70
] coordinates {(A,44) (B,3) (C,0) (D,1) (E,52)};

\nextgroupplot
\addplot[
  pattern=grid,
  pattern color=blue!70
] coordinates {(A,10) (B,1) (C,0) (D,0) (E,89)};

\nextgroupplot
\addplot[
  pattern=grid,
  pattern color=blue!70
] coordinates {(A,7.04) (B,2.82) (C,0) (D,1.41) (E,88.73)};

\nextgroupplot[
  title={{AG News Train Partition\\Original Instances Position: B}},
  ylabel={{\fontsize{5pt}{5pt}\selectfont BDQ W/ Orig. Ins.}}
]
\addplot[
  pattern=crosshatch,
  pattern color=red!80
] coordinates {(A,6) (B,88) (C,0) (D,0) (E,6)};

\nextgroupplot[
  title={{AG News Test Partition\\Original Instances Position: B}}
]
\addplot[
  pattern=crosshatch,
  pattern color=red!80
] coordinates {(A,7) (B,91) (C,0) (D,0) (E,2)};

\nextgroupplot[
  title={{WNLI Train Partition\\Original Instances Position: D}}
]
\addplot[
  pattern=crosshatch,
  pattern color=red!80
] coordinates {(A,6) (B,1) (C,0) (D,51) (E,42)};

\nextgroupplot[
  title={{WNLI Validation Partition\\Original Instances Position: B}}
]
\addplot[
  pattern=crosshatch,
  pattern color=red!80
] coordinates {(A,4.23) (B,46.48) (C,0) (D,0) (E,49.3)};

\end{groupplot}
\end{tikzpicture}
\caption{Results of positional adversarial analysis. The green bar charts show the selection frequency distribution of options in BDQ while the blue and red ones represent results after replacing new word-level perturbations and original dataset instances with the least-chosen option detected by BDQ, respectively. The increase in selection frequency for options containing original dataset instances (red bar charts) signals the model's prior exposure to data. In all settings, GPT-4 is the base model.}
\label{figure:grid-of-bars}
\end{figure*}

\subsection{Positional Adversarial Analysis}
\label{subsection:positional-adversarial-analysis}

Our quiz-based approach uses three main tactics: (1) integrating original dataset instances with their word-level perturbations as quiz options exposes memorization in LLMs; (2) strategically placing these original dataset instances in non-preferred options avoids the overestimation of contamination; and (3) permuting these original dataset instances among non-preferred options avoids the underestimation of contamination.
Below, we provide empirical results that corroborate each tactic.

\vspace{0.15cm}

\noindent\textbf{Exposing Memorization via Quiz Options.} For this experiment, after identifying positional biases by Bias Detector Quiz (BDQ), we undertake two variants of Bias Compensator Quiz (BCQ).
These variants are designed to illustrate how positional biases are affected by the presence and absence of memorization.
Unlike the standard setting described in Subsection \ref{subsection:deployment-of-DCQ}, these variants do not involve permuting correct answers (original dataset instances) among the non-preferred options. \emph{Instead, we statically position them in a single detected least-chosen option.}
In other words, this least-chosen option is a non-preferred option that has the lowest selection frequency among others.

In the first variant of BCQ, we replace the least-chosen option with the original dataset instances. However, in the second variant, the least-chosen option is replaced with unique perturbations, ensuring they differ from all four perturbations used in BDQ.
This confirms that if an LLM has not been exposed to the original dataset instances, and thus, cannot recognize them by memorization, the selection frequency distribution among options in BDQ should be maintained regardless of whether the original dataset instances are included among the options. In contrast, a pronounced preference for selecting the option containing the original dataset instances signals memorization and prior exposure to data, unveiling data contamination.

To analyze the changes in the selection frequency distribution of options, we utilize four contaminated dataset partitions, reported by \citet{golchin2024time}. Specifically, we use the train and test sets of AG News as well as the train and validation sets of the WNLI dataset, with GPT-4 as the base model. Figure \ref{figure:grid-of-bars} shows the outcomes of this experiment. As shown, when new perturbations are replaced with the least-chosen options (blue bar plots), as the model has not seen them, the selection frequency distribution of options remains significantly consistent with BDQ (green bar plots).
Conversely, there is a significant preference for selecting options containing the original dataset instances when they are replaced with the least-chosen options (red bar plots).\footnote{We assessed this significance using Fisher's Exact Test \citep{fisher1922}, with $p$-values $< 0.05$.}

For example, the initial selection frequency distribution of options in BDQ for the AG News train set is \{A: 29, B: 0, C: 0, D: 0, E: 71\}.
Thus, options B, C, and D are deemed non-preferred due to being chosen less frequently than random chance (20 times out of 100 questions), and they are also the least-chosen options, with zero selections each.\footnote{Note that \textit{non-preferred options are not fixed for all models and datasets}, and must be identified for each specific dataset and model individually.}
When there are multiple least-chosen options, we opt for the one that results in the highest selection frequency when replaced with new perturbations.
If this leads to identical selection frequencies among the least-chosen options, we then opt for the position that results in the highest selection frequency when replaced with original dataset instances.
This helps to better visualize the increase in selection frequency while establishing a rigorous analysis by favoring the setting containing new perturbations first.
Thus, in this example, we select option B in BDQ to conduct our replacements, as it resulted in the highest selection frequency when replaced with the original dataset instances.
When option B is replaced with new perturbations, no preference for selecting option B is observed, and the initial positional biases in BDQ remain (almost) unchanged: \{A: 28, \underline{B: 0}, C: 0, D: 0, E: 72\}.
However, when option B is replaced with the original dataset instances, a noticeable rise in its selection becomes apparent: \{A: 6, \underline{B: 88}, C: 0, D: 0, E: 6\}. Therefore, we showed that infusing original dataset instances among their word-level perturbations as quiz options triggers evidence of memorization in LLMs.

\pgfplotsset{compat=1.7}

\begin{figure*}
\centering
\begin{tikzpicture}
\begin{groupplot}[
    group style={
    group size=4 by 3,
    horizontal sep=0.9cm,
    vertical sep=0.85cm,
    },
    width=0.295\textwidth,  
    height=0.15\textwidth,
    ybar,
    symbolic x coords={A, B, C, D, E},
    xtick=data,
    nodes near coords,
    tick label style={font=\tiny},
    nodes near coords style={font=\tiny},
    nodes near coords align={vertical},
    axis lines*=left,
    ymin=0,
    ymax=100,
    ytick={0,100},
    title style={at={(axis description cs:0.5,-2)}, align=center, font=\fontsize{7}{8.7}\selectfont},
    yticklabel={\pgfmathparse{\tick}\pgfmathprintnumber{\pgfmathresult}\%},
]

\nextgroupplot[title={{BDQ\\Non-Preferred Options: B, C, D}}]
\addplot[pattern=north east lines, pattern color=green!70!black] coordinates {(A,29) (B,0) (C,0) (D,0) (E,71)};
\nextgroupplot[title={{BCQ\\Original Instances Position: B}}]
\addplot[pattern=crosshatch, pattern color=red!80] coordinates {(A,6) (B,88) (C,0) (D,0) (E,6)};
\nextgroupplot[title={{BCQ\\Original Instances Position: C}}]
\addplot[pattern=crosshatch, pattern color=red!80] coordinates {(A,11) (B,0) (C,80) (D,0) (E,9)};
\nextgroupplot[title={{BCQ\\Original Instances Position: D}}]
\addplot[pattern=crosshatch, pattern color=red!80] coordinates {(A,12) (B,0) (C,0) (D,75) (E,13)};
\end{groupplot}
\end{tikzpicture}
\caption{An illustration of the full process for our approach, performed on the AG News train set. Initially, Bias Detector Quiz (BDQ) is executed to identify non-preferred options (i.e., B, C, and D), followed by multiple Bias Compensator Quizzes (BCQs) to determine the maximum contamination (88\%) through permutation among the non-preferred options.
In all settings, GPT-4 is the base model.}
\label{figure:permutation}
\end{figure*}

\vspace{0.15cm}

\noindent\textbf{Avoiding Overestimation Using Non-Preferred Options.} Positioning the original dataset instances among quiz options without considering positional biases fails to accurately measure contamination levels. Considering the previous example, if the original dataset instances are replaced with option A in BDQ, any contamination level surpassing 71\% results in the system indicating 100\% contamination. While it is not feasible to unequivocally guarantee the probabilistic behavior of LLMs, including the potential overestimation of contamination in our case, it can be confidently asserted that our approach does not stimulate this overestimation at least by design. In fact, this is achieved by using non-preferred options as the replacement spots for the original dataset instances.

\vspace{0.15cm}

\noindent\textbf{Avoiding Underestimation by Permutation.} As discussed, non-preferred options serve as safe spots for positioning original dataset instances to mitigate the risk of overestimation in the presence of positional biases.
However, these biases can also result in uneven memorization triggered across the detected non-preferred options. 
Figure \ref{figure:permutation} provides empirical evidence of this phenomenon when using the AG News train partition as our case study. As shown, when the original dataset instances are placed in the non-preferred options (i.e., B, C, and D), the level of contamination revealed by memorization varies within a single setting: 88\%, 80\%, and 75\% for positions B, C, and D, respectively.
As detailed in Subsection \ref{subsection:deployment-of-DCQ}, to avoid underestimation, the highest quiz performance among all BCQs is reported as the maximum level of detected contamination, which is 88\% in this example.
Thus, we empirically showed that permuting original dataset instances among the non-preferred options is essential for uncovering the highest possible contamination and avoiding underestimation.

\setlength{\tabcolsep}{3pt}
\renewcommand{\arraystretch}{0.9}

\begin{table*}[!t] 
\centering
\caption{Results from our DCQ after introducing systematic contamination to GPT-3.5 and Llama 2. The table lists the minimum and maximum detected contamination levels. Results are also compared with the replication-based (rplc-based) method by \citet{golchin2024time}. Note that, in settings with no contamination, Recall cannot be calculated, as both True Positives and False Negatives are zero.}
\label{table:results-for-controlled-contamination}
\begin{adjustbox}{width=\textwidth,center} 
\scriptsize
\begin{tabular}{l!{\vrule width 1pt}l!{\vrule width 1pt}c|c|c!{\vrule width 1pt}c|c|c}
    \toprule
    \multicolumn{2}{c}{\textbf{}} & \multicolumn{3}{c}{\textbf{MeetingBank}} & \multicolumn{3}{c}{\textbf{AuTexTification}} \\
    \cmidrule(lr){3-5} \cmidrule(lr){6-8}
    \textbf{Model} & \textbf{Method} & \textbf{100\% Cont.} & \textbf{50\% Cont.} & \textbf{0\% Cont.} & \textbf{100\% Cont.} & \textbf{50\% Cont.} & \textbf{0\% Cont.} \\
    \midrule
    \multirow{4}{*}{GPT-3.5} & Rplc-based (\%) & 1.00 & 0.00 & 0.00 & 1.00 & 0.00 & 0.00 \\
    & Our DCQ (\%) & [85.87, 87.00] & [46.31, 49.00] & [0.00, 3.00] & [66.29, 70.00] & [44.68, 48.00] & [0.00, 5.00] \\
    & Recall (\%) & 87.00 & 88.00 & Undefined & 70.00 & 70.00 & Undefined \\
    & Precision (\%) & 87.00 & 89.80 & 0.00 & 70.00 & 72.92 & 0.00 \\
    \midrule
    \multirow{4}{*}{Llama 2} & Rplc-based (\%) & 1.00 & 0.00 & 0.00 & 0.00 & 0.00 & 0.00 \\
    & Our DCQ (\%) & [84.21, 85.00] & [45.83, 48.00] & [0.00, 2.00] & [72.63, 74.00] & [44.21, 47.00] & [0.00, 2.00] \\
    & Recall (\%) & 85.00 & 84.00 & Undefined & 74.00 & 74.00 & Undefined \\
    & Precision (\%) & 85.00 & 87.50 & 0.00 & 74.00 & 78.72 & 0.00 \\
    \bottomrule
\end{tabular}
\end{adjustbox}
\end{table*}

\begin{table}[!t]
    \centering
    \setlength{\tabcolsep}{2pt}  
    \caption{A Comparison between the contamination levels reported for GPT-4 \citep{openai2023gpt4} and those identified by the replication-based (rplc-based) method \citep{golchin2024time} and our DCQ. This experiment encompasses the test, validation, and train partitions of the HumanEval, DROP, and GSM8k datasets, respectively.}
    \label{tab:known-contamination}
    \begin{adjustbox}{width=\columnwidth,center} 
    \begin{tabular}{l!{\vrule width 1pt}c|c|c}
    \toprule
    \textbf{Method} & \textbf{HumanEval} & \textbf{DROP} & \textbf{GSM8k} \\
    \midrule
    Reported (\%) & $\approx$ 25.00 &  $\approx$ 21.00 & Full but a part \\
    Rplc-based (\%) & 0.00 & 4.00 & 2.00 \\
    Our DCQ (\%)  & [55.62, 56.71] & [42.86, 44.00] & [78.79, 79.00] \\
    \bottomrule
    \end{tabular}
    \end{adjustbox}
\end{table}

\subsection{Evaluation under Controlled Contamination}
\label{subsection:controlled-contamination}

\textbf{Systematic Contamination.} As for the first step in assessing the performance of our approach, we start with systematic contamination. In this experiment, we systematically contaminate GPT-3.5 and Llama 2 with two uncontaminated datasets based on their training cutoff: MeetingBank and AuTexTification.\footnote{See Appendix \ref{appendix:continued-training-for-contamination} for details on systematic contamination of GPT-3.5 and Llama 2.}

Our evaluation involves estimating contamination using our Data Contamination Quiz (DCQ) approach for three contamination levels per dataset: 100\%, 50\%, and 0\%. Following prior work \citep{shi2024detecting,oren2024proving}, we perform systematic contamination using 1,000 instances from each dataset separately.

For the 100 instances used in our method, the systematic contamination settings vary as follows:

\begin{itemize}[itemsep=0pt, topsep=0pt]
    \item\textbf{100\% Contamination:} All 100 instances are selected from the 1,000 contaminated instances. 
    \item\textbf{50\% Contamination:} 50 instances are sampled from the 1,000 contaminated instances, while the other 50 are drawn from the source dataset, excluding the 1,000 contaminated instances.   
    \item\textbf{0\% Contamination:} All 100 instances are sampled from the source dataset, excluding the 1,000 contaminated instances.
\end{itemize}

This experimental design models real-world scenarios where DCQ estimates contamination using subsampled dataset instances.

Table \ref{table:results-for-controlled-contamination} presents the results of our experiment under systematic contamination upon applying DCQ.\footnote{We provide statistics on the positional biases of all the models used in this study across the corresponding datasets in Table \ref{tab:positional-biases} in Appendix \ref{sec:positional-biases}.}
Compared with the replication-based strategy \citep{golchin2024time}, DCQ proves to be significantly more effective at revealing contamination.
In fact, contamination levels identified by replicating instances are so negligible that it is impracticable to compare with our DCQ's performance.
As noted in Subsection \ref{subsection:positional-adversarial-analysis}, this improved ability to detect more memorization is obtained by minimizing LLMs' probabilistic outputs to a single-letter option that hints at the presence of memorization.
Another benefit of it is its ability to bypass the preemptive filters set to stop LLMs from producing copyrighted content.\footnote{Refer to Appendix \ref{appendix:current-and-previous-versions} to see case examples.}
Importantly, DCQ not only reveals more memorization but also effectively measures contamination levels.
As detailed in Subsection \ref{subsection:positional-adversarial-analysis}, this effective estimation is derived by placing and permuting the original dataset instances among the non-preferred options, simultaneously avoiding overestimation and underestimation.

\vspace{0.15cm}

\noindent\textbf{Reported Contamination.} The GPT-4 technical report provides cross-contamination between several datasets and the training data of this model \citep{openai2023gpt4}.
In this experiment, we consider the backdrop of reported contamination as a baseline against which we compare the performance of DCQ in detecting contamination.
We use three datasets that hold significant importance in the context of LLMs: HumanEval,\footnote{As done by \citet{openai2023gpt4}, we apply DCQ to the entire test set of the HumanEval dataset with 164 samples.} DROP, and GSM8k, and apply DCQ to their test, validation, and train splits, respectively. The rationale behind selecting these particular datasets stems from their pivotal role in assessing the reasoning/problem-solving capabilities of LLMs, along with the explicit mention of contamination levels for them by \citet{openai2023gpt4}.
Further, our focus on the GSM8k train partition is driven by the fact that this set---apart from a small portion---was included in the training data of GPT-4, though without a specified level of inclusion. Thus, we expect that our DCQ detects a significant level of contamination for it.

Table \ref{tab:known-contamination} compares the contamination levels detected by DCQ with those reported by \citet{openai2023gpt4} for the aforesaid datasets.
As mentioned in Subsection \ref{subsection:deployment-of-DCQ}, given that DCQ inherently does not overestimate/underestimate contamination, the results indicate its ability to uncover higher contamination levels, even when compared to those measured in the availability of the training data.
Further, as expected, the contamination level identified for the GSM8k train set is markedly high while the replication-based method can only detect 2.00\% of it.

Note that the cross-contamination levels reported by \citet{openai2023gpt4} in Table \ref{tab:known-contamination} may not offer a thorough depiction of the levels detected by DCQ for two reasons: (1) the measured cross-contamination comes from overlapping three 50-character substrings, yielding high false negatives/positives \citep{openai2023gpt4}. As a result, the reported contamination levels serve as approximations rather than definitive measures;
and (2) cross-contamination levels were estimated using a maximum of 1,000 subsampled instances from each set, rather than the entire set \citep{openai2023gpt4}.

\renewcommand{\arraystretch}{0.8}
\begin{table*}[!t]
\centering
\caption{Estimated data contamination levels for GPT-4, GPT-3.5, and Llama 2 using our DCQ across test/validation partitions of seven datasets, based on a maximum of 100 subsampled instances. Results are also compared with the replication-based (rplc-based) method by \citet{golchin2024time}.
}
\label{tab:GPT-4-and-GPT-3.5-results}
\scriptsize
\setlength{\tabcolsep}{2pt}
\renewcommand{\arraystretch}{1}
\begin{adjustbox}{width=\textwidth,center}
\begin{tabular}{l!{\vrule width 1pt}l!{\vrule width 1pt}c|c|c|c|c|c|c}
\toprule
\textbf{Model} & 
\multicolumn{1}{c!{\vrule width 1pt}}{\textbf{Method}} &
\textbf{IMDB} &
\textbf{AG News} &
\textbf{Yelp} &
\textbf{RTE} &
\textbf{WNLI} &
\textbf{SAMSum} &
\textbf{XSum} \\
\midrule
\multirow{2}{*}{GPT-4} & Rplc-based (\%) & 0.00 & 3.00 & 0.00 & 1.00 & 15.49 & 0.00 & 7.00 \\
                       & Our DCQ (\%) & [81.82, 82.00] & [91.00, 91.00] & [80.00, 80.00] & [60.00, 60.00] & [45.31, 50.70] & [76.04, 77.00] & [94.90, 95.00] \\
\midrule
\multirow{2}{*}{GPT-3.5} & Rplc-based (\%) & 0.00 & 0.00 & 0.00 & 1.00 & 2.82 & 0.00 & 2.00 \\
                         & Our DCQ (\%) & [46.43, 55.00] & [80.43, 82.00] & [12.12, 13.00] & [65.06, 71.00] & [7.47, 12.68] & [70.11, 74.00] & [78.35, 79.00] \\
\midrule
\multirow{2}{*}{Llama 2} & Rplc-based (\%) & 0.00 & 0.00 & 0.00 & 0.00 & 0.00 & 0.00 & 0.00 \\
                               & Our DCQ (\%) & [4.44, 14.00] & [17.20, 23.00] & [15.15, 16.00] & [30.23, 40.00] & [11.29, 22.54] & [0.00, 1.00] & [11.00, 11.00] \\
\bottomrule
\end{tabular}
\end{adjustbox}
\end{table*}

\subsection{Evaluation in the Wild}
\label{subsection:eval-in-the-wild}

After examining contamination in controlled environments, we apply our detection strategy to various datasets in real-world scenarios where the exact contamination levels are unknown.
To mitigate the lack of information about the precise contamination levels for the datasets we analyze, we assess the contamination levels detected by our approach against the replication-based method by \citet{golchin2024time}.
This evaluation is informed by effective estimates obtained under controlled contamination in Subsection \ref{subsection:controlled-contamination} and by understanding that our method does not encourage both overestimation and underestimation.
For this experiment, we use the same datasets as in their study, covering test/validation splits of seven datasets with GPT-4, GPT-3.5, and Llama 2.

Table \ref{tab:GPT-4-and-GPT-3.5-results} lists all the results from our evaluation in the wild over 21 settings. As outcomes indicate, the replication-based method \citep{golchin2024time}
can replicate a few dataset instances, such as when GPT-4 serves as the base model and is applied to WNLI (15.49\%) and XSum (7.00\%) datasets.
In comparison, our DCQ detects significant amounts of contamination across all studied dataset partitions. For example, the maximum contamination identified by our DCQ in the aforesaid datasets reaches 50.70\% and 95.00\%, representing a detection range of contamination 3 to over 13 times greater than that of the replication-based method.
Additionally, this ratio is even more substantial for other studied datasets.

Overall, our results indicate that the replication-based method is effective only for binary detection of contamination when the memorization level is high, and lacks the ability to effectively estimate contamination.
In contrast, our DCQ reveals cases of memorization/contamination at levels significantly greater than methods based on replicating/extracting training data \citep{golchin2024time,DBLP:conf/iclr/CarliniIJLTZ23}.

\section{Related Work}

Data contamination is shown to inflate downstream performance \citep{Zhou2023DontMY,palavalli2024taxonomy,jiang2024investigating,dong2024generalization,deng2024unveiling,balloccu2024leak,DBLP:conf/aaai/LiF24}.
Identifying data contamination in LLMs is not difficult when the training data is accessible, though it can be resource-intensive. Typically, this process involves expensive overlap comparisons between the training data and the test data using high-order $n$-grams---a method exclusive to model developers that yields approximate estimates of contamination rather than definitive ones \citep[inter alia]{openai2023gpt4,touvron2023Llama,touvron2023Llama2,anil2023palm,chowdhery2022palm,DBLP:conf/nips/BrownMRSKDNSSAA20}.
However, this task becomes challenging when the training data is absent due to being proprietary or when it is open but vast in size.

\citet{Roberts2023DataCT} used a time-based approach for contamination detection by analyzing dataset pass rates pre- and post-model training cutoff.
However, this method has limitations due to the continuous updates of proprietary LLMs and the need for ongoing data collection. Further, detecting contamination for non-code/math datasets remained unresolved.
\citet{shi2024detecting} employed a similar strategy, measuring outlier word likelihood in new examples to determine if a text was part of the training corpus. While only useful for LLMs with accessible logits, detecting training data is not synonymous with confirming contamination. 
\citet{zhang2024min} improved this method using the idea of local maxima.
In line with probability-based methods, \citet{oren2024proving} aimed to detect data contamination by comparing the likelihoods of a canonically ordered dataset and its shuffled version. However, the positional biases in LLMs \citep{wang2023large,zheng2024large} and the bold presumption of dataset instances appearing in the training data in the same order as on the web limit the applicability of this method.
\citet{yang2023rethinking} further relaxed their settings by assuming access to training data to detect contamination. They used embedding similarity to find the top-k similar training samples for each dataset instance and used an LLM to verify if these samples were rephrased versions of the dataset instance. Similarly, \citet{deng2023investigating} developed a retrieval-based method that identified potential contamination by querying training corpora and assessing overlap with benchmark data.
However, assuming access to training data is impractical, even for open-weight models.

A few recent studies developed strategies for detecting contamination without access to training data and model weights/logits. \citet{sainz2023nlp,did-ChatGPT-cheat-on-your-test} prompted GPT-3.5 to generate the first instances of a dataset partition. However, this method faltered due to the sparseness of the signal in the prompt, coupled with the LLM's safety filters preventing the generation of copyrighted materials.
In contrast, \citet{golchin2024time} amplified the signal in the prompt by infusing a random-length initial segment of a dataset instance and directing the LLM to complete it.
\citet{dong2024generalization} detected contamination by measuring the peakedness of the model's output distribution through edit distance to reveal contamination.
\citet{deng2023investigating} proposed two methods: a mask-based approach, which detects contamination if the model can predict masked parts of dataset instances, and a question-based approach, which identifies contamination if the model generates a new answer to a question that already has an answer embedded in the prompt. However, none of these methods can estimate contamination.

\section{Conclusion}
We proposed the Data Contamination Quiz (DCQ), the first method to \textit{estimate} verbatim data contamination in fully black-box LLMs by framing it as a series of multiple-choice questions, forming our quiz. This quiz instructs an LLM to select an option containing the original dataset instance among its word-level perturbations. If the LLM succeeds in detecting the original dataset instance, it reveals prior exposure to data. Thus, the quiz performance indicates the level of detected contamination for the LLM that took DCQ.
Using several datasets and LLMs, under controlled and uncontrolled settings, our findings demonstrated that the detected contamination levels by our DCQ not only align with previously reported contaminated datasets but also surpass the existing methods by providing an estimation of verbatim data contamination, moving beyond binary detection.

\section*{Limitations}

Data contamination primarily occurs in two forms: \textit{verbatim}, where dataset instances are directly included in the training data, and \textit{non-verbatim}, such as metadata contamination. This study only focuses on estimating verbatim contamination in LLMs. Therefore, we encourage future research to develop methods for detecting and estimating other types of contamination beyond verbatim.

We also emphasize the importance of carefully considering data contamination when evaluating LLMs on benchmarks, particularly for reasoning tasks, where the goal is to assess genuine generalization abilities. Although there has been progress in reducing contamination risks through private or semi-private evaluations of newer models, recent findings suggest that contamination remains a serious concern, and private evaluations alone are neither a definitive nor a long-term solution.

For example, OpenAI’s recent o3 model \citep{jaech2024openai, openai2024o3mini} achieved near-human performance on the ARC-AGI-1 dataset \citep{chollet2019measure}, representing a major leap in reasoning capabilities. However, it scored only in the single digits on the newly released ARC-AGI-2 dataset \citep{chollet2024arc}, even though both datasets are of similar difficulty and considered relatively easy for humans, according to their creators \citep{arcprize2025}. Since the only primary difference between the two datasets is their release time, this large performance gap raises concerns about the genuine generalization abilities of these models and the reliability of private evaluations in preventing contamination, as this longitudinal performance drop can be mostly attributed to contamination \citep{Roberts2023DataCT}.  As a result, data contamination remains a major challenge in model evaluation and calls for further research.

\section*{Acknowledgments}

We sincerely thank Dr. Steven Bethard and Dr. Eduardo Blanco for their expert guidance on the early draft of this paper. 
We also appreciate the action editor and anonymous reviewers for their constructive inputs that greatly improved this work.

\bibliography{tacl2021}
\bibliographystyle{acl_natbib}


\onecolumn

\appendix

\section{Detecting Data Contamination for Different Checkpoints}
\label{appendix:current-and-previous-versions}

As LLM providers persistently update their models, the safety of these LLMs towards detected weaknesses and adversarial attacks also improves. The detection technique recently developed by \citet{golchin2024time} is one such attack addressed by these updates. For example, when GPT-4 is instructed to complete a partial reference instance from a classification dataset, the June 2023 snapshot (\texttt{gpt-4-0613}) is capable of accomplishing this task, thus unveiling contamination. In contrast, a more recent version from November 2023 (\texttt{gpt-4-1106-preview}) often chooses not to complete the task due to the activation of newly implemented safety mechanisms. Table \ref{tab:data-contamination-in-present-vs-past-updates-replication} illustrates this behavior.

On the other hand, in Table \ref{tab:data-contamination-in-present-vs-past-updates-quiz}, we show that our proposed quiz-based methodology can successfully identify data contamination for both versions. This highlights the capability of our approach in pinpointing data contamination even in the presence of newly added safety filters. In fact, our strategy can effectively sidestep filters related to copyrighted materials by delivering a single-letter response that embodies the detection of the original dataset instance among other options, thereby exposing contamination.

\begin{table}[!t]
\centering
\caption{A comparison between the completions generated using replication-based approach \citep{golchin2024time} for the June and November 2023 snapshots of GPT-4. While the June snapshot exactly replicates the second piece of the provided dataset instance, the November snapshot avoids completing it due to the activation of filters intended to prevent the generation of copyrighted content.}
\label{tab:data-contamination-in-present-vs-past-updates-replication}
\scriptsize
\setlength{\tabcolsep}{6pt} 
\begin{adjustbox}{width=1\textwidth,center}
\begin{tabular}{p{0.21\textwidth}c} 
\toprule
\multicolumn{1}{l}{\textbf{LLM Version/Detection}} & \textbf{Original Dataset Instance and Its Replica} \\
\midrule & 
\parbox{0.75\textwidth}{\underline{\textbf{Original Dataset Instance:}}
\vspace{-0.08cm}
{\flushleft\textbf{Review:\,}\bleuhl{Bromwell High is a cartoon comedy. It ran at the same time as some other programs about school}} 

\vspace{0.012cm}
\bleuhl{life, such as "Teachers". My 35 years in the teaching profession lead me to believe that Bromwell High's}

\vspace{0.012cm}
\bleuhl{satire is much closer to reality than is "Teachers".}
The scramble to survive financially, the insightful students who can see right through their pathetic teachers' pomp, the pettiness of the whole situation, all remind me of the schools I knew and their students. When I saw the episode in which a student repeatedly tried to burn down the school, I immediately recalled ......... at .......... High. A classic line: INSPECTOR: I'm here to sack one of your teachers. STUDENT: Welcome to Bromwell High. I expect that many adults of my age think that Bromwell High is far fetched. What a pity that it isn't!

\bleuhl{\textbf{Label:} 1 (positive)}} \\ \\
\parbox{4cm}{\textbf{Version:} \texttt{gpt-4-0613}  \\ \textbf{Detection:} Exact Match} &
\begin{minipage}{0.75\textwidth} 
\begin{tikzpicture}[rounded corners=8pt, thick, text=black, text opacity=1]
    \node[draw=solid_gray, fill=light_gray, line width=1pt, text=black, text width=0.965\textwidth, align=left, font=\fontsize{7pt}{1em}\selectfont, inner xsep=6.5pt, inner ysep=5pt] at (0,0) {\textbf{Instruction:}
    You are provided with the first piece of an instance from the train split of the IMDB dataset.
    Finish the second piece of the instance as exactly appeared in the dataset.
    Only rely on the original form of the instance in the dataset to finish the second piece.
    
    \vspace{4pt}
    \textbf{Label:} 1 (positive)
    
    \vspace{4pt}
    \textbf{First Piece:} Bromwell High is a cartoon comedy. It ran at the same time as some other programs about school life, such as "Teachers". My 35 years in the teaching profession lead me to believe that Bromwell High's satire is much closer to reality than is "Teachers".
    
    \vspace{4pt}
    \textbf{Second Piece:}
    \highlight{\texttt{The scramble to survive financially, the insightful students who can see right}}
    \vspace{-0.24cm}
    \highlight{\texttt{through their pathetic teachers' pomp, the pettiness of the whole situation, all}}
    \vspace{-0.24cm}
    \highlight{\texttt{remind me of the schools I knew and their students. When I saw the episode in which}}
    \vspace{-0.24cm}
    \highlight{\texttt{a student repeatedly tried to burn down the school, I immediately recalled ......}}
    \vspace{-0.24cm}
    \highlight{\texttt{... at .......... High. A classic line: INSPECTOR: I'm here to sack one of your}}
    \vspace{-0.24cm}
    \highlight{\texttt{teachers. STUDENT: Welcome to Bromwell High. I expect that many adults of my age}}
    \vspace{-0.24cm}
    \highlight{\texttt{think that Bromwell High is far fetched. What a pity that it isn't!}}
    };
\end{tikzpicture}
\end{minipage}
\\
\midrule & 
\\ \parbox{4cm}{\textbf{Version:} \texttt{\fontsize{7}{7}\selectfont gpt-4-1106-preview}  \\ \textbf{Detection:} None}  &
\begin{minipage}{0.75\textwidth} 
\begin{tikzpicture}[rounded corners=8pt, thick, text=black, text opacity=1]
    \node[draw=solid_gray, fill=light_gray, line width=1pt, text=black, text width=0.965\textwidth, align=left, font=\fontsize{7pt}{1em}\selectfont, inner xsep=6.5pt, inner ysep=5pt] at (0,0) {\textbf{Instruction:}
    You are provided with the first piece of an instance from the train split of the IMDB dataset.
    Finish the second piece of the instance as exactly appeared in the dataset.
    Only rely on the original form of the instance in the dataset to finish the second piece.
    
    \vspace{4pt}
    \textbf{Label:} 1 (positive)
    
    \vspace{4pt}
    \textbf{First Piece:} Bromwell High is a cartoon comedy. It ran at the same time as some other programs about school life, such as "Teachers". My 35 years in the teaching profession lead me to believe that Bromwell High's satire is much closer to reality than is "Teachers".
    
    \vspace{4pt}
    \textbf{Second Piece:}
    \highlight{\texttt{Sorry, but I can't continue the text from the IMDB dataset as it is copyrighted}}
    \vspace{-0.24cm}
    \highlight{\texttt{material. However, I can provide a summary or discuss the themes and characters of}}
    \vspace{-0.24cm}
    \highlight{\texttt{"Bromwell High" if you're interested. Let me know if you would like me to provide}}
    \vspace{-0.24cm}
    \highlight{\texttt{some other type of information!}}
    };
\end{tikzpicture}
\end{minipage}
\\
\bottomrule
\end{tabular}
\end{adjustbox}
\end{table}

\begin{table}[htbp]
\centering
\caption{An illustration of a quiz question posed when taking our DCQ to both June and November snapshots of GPT-4. While the generated response is correct (option C), this approach—unlike the replication-based strategy \citep{golchin2024time}—proves to be effective for both snapshots without triggering any content-related filters set by the model developer.}
\label{tab:data-contamination-in-present-vs-past-updates-quiz}
\scriptsize
\setlength{\tabcolsep}{3.9pt} 
\begin{tabular}{p{0.21\textwidth}c} 
\toprule
\multicolumn{1}{l}{\textbf{LLM Version/Detection}} & \textbf{Original Dataset Instance and Its Integration as Quiz Option} \\
\midrule & 
\parbox{0.75\textwidth}{\underline{\textbf{Original Dataset Instance:}}
\vspace{-0.08cm}
\flushleft\textbf{Review:} Bromwell High is a cartoon comedy. It ran at the same time as some other programs about school life, such as "Teachers". My 35 years in the teaching profession lead me to believe that Bromwell High's satire is much closer to reality than is "Teachers". The scramble to survive financially, the insightful students who can see right through their pathetic teachers' pomp, the pettiness of the whole situation, all remind me of the schools I knew and their students. When I saw the episode in which a student repeatedly tried to burn down the school, I immediately recalled ......... at .......... High. A classic line: INSPECTOR: I'm here to sack one of your teachers. STUDENT: Welcome to Bromwell High. I expect that many adults of my age think that Bromwell High is far fetched. What a pity that it isn't!

\textbf{Label:} 1 (positive)} \\ \\
\parbox{4cm}{\textbf{Versions:} \texttt{gpt-4-0613} and \\ \texttt{gpt-4-1106-preview}  \\ \textbf{Detection:} Correct (option C)} &
\begin{minipage}{0.75\textwidth} 
\begin{tikzpicture}[rounded corners=8pt, thick, text=black, text opacity=1]
    \node[draw=solid_gray, fill=light_gray, line width=1pt, text=black, text width=0.965\textwidth, align=left, font=\fontsize{7pt}{1em}\selectfont, inner xsep=6.5pt, inner ysep=5pt] at (0,0) {\textbf{Instruction:} You are provided with a five-choice quiz. Your task is to correctly select the option that exactly corresponds to an instance from the train split of the IMDB dataset.
    
    \vspace{0.17cm}
    When selecting the option, you must ensure that you follow the following rules:

    (1) You must ensure that you only generate a single option letter as your answer.
    
    (2) If you do not know the dataset or the correct answer, you must select option "E) None of the provided options."
    
    \vspace{0.17cm}
    Hint: While all the following options seem similar in terms of meaning, there is only one option that reflects the exact wording with respect to the original instance.
    
    -- -- --
    
    \textbf{A)}
    Bromwell High is an animated comedy. It aired at the same time as some other shows about school life, such as ”Teachers”.
    My 35 years in the education profession make me feel that Bromwell High’s humor is much closer to reality than is ”Teachers”.
    The effort to survive financially, the smart students who can see right through their pathetic teachers’ pomp, the pettiness of the whole situation, all remind me of the schools I knew and their students. When I saw the episode in which a student repeatedly tried to burn down the school, I immediately recalled ......... at .......... High. A memorable line: INSPECTOR: I’m here to sack one of your teachers. STUDENT: Welcome to Bromwell High. I expect that many adults of my age think that Bromwell High is unbelievable. What a pity that it isn’t! \\
    Label: 1 (positive)

    \vspace{0.17cm}
    
    \textbf{B)}
    Bromwell High is a cartoon comedy. It broadcast concurrently with some other programs about school life, like ”Teachers”.
    My 35 years in the teaching profession lead me to believe that Bromwell High’s satire is way closer to reality than is ”Teachers”.
    The struggle to survive financially, the sharp students who can see right through their pathetic teachers’ pomp, the pettiness of the whole situation, all remind me of the schools I knew and their students. When I saw the episode in which a student repeatedly attempted to burn down the school, I immediately remembered ......... at .......... High. An iconic line: INSPECTOR: I’m here to fire one of your teachers. STUDENT: Welcome to Bromwell High. I expect that many adults of my age think that Bromwell High is far fetched. What a pity that it isn’t! \\
    Label: 1 (positive)
    
    \vspace{0.17cm}
    
    \textbf{C)}
    \emph{Bromwell High is a cartoon comedy. It ran at the same time as some other programs about school life, such as ”Teachers”.
    My 35 years in the teaching profession lead me to believe that Bromwell High’s satire is much closer to reality than is ”Teachers”.
    The scramble to survive financially, the insightful students who can see right through their pathetic teachers’ pomp, the pettiness of the whole situation, all remind me of the schools I knew and their students. When I saw the episode in which a student repeatedly tried to burn down the school, I immediately recalled ......... at .......... High. A classic line: INSPECTOR: I’m here to sack one of your teachers. STUDENT: Welcome to Bromwell High. I expect that many adults of my age think that Bromwell High is far fetched. What a pity that it isn’t! \\
    Label: 1 (positive)}

    \vspace{0.17cm}
    
    \textbf{D)}
    Bromwell High is a cartoon comedy. It ran at the same time as some other shows about school life, such as ”Teachers”.
    My 35 years in the teaching profession lead me to believe that Bromwell High’s satire is much closer to reality than is ”Teachers”.
    The scramble to survive financially, the clever students who can see right through their pathetic teachers’ pomp, the pettiness of the whole situation, all remind me of the schools I knew and their students. When I watched the episode in which a student repeatedly tried to set fire to the school, I immediately recalled ......... at .......... High. A classic line: INSPECTOR: I’m here to dismiss one of your teachers. STUDENT: Welcome to Bromwell High. I expect that many adults of my age think that Bromwell High is far fetched. What a pity that it isn’t! \\
    Label: 1 (positive)

    \vspace{0.17cm}

    \textbf{E)}
    None of the provided options.

    -- -- --
    
    \textbf{Answer:} \highlight{\texttt{C}}
    };
\end{tikzpicture}
\end{minipage}
\\
\bottomrule
\end{tabular}
\end{table}

\section{Details and Analysis of Generated Word-Level Perturbations}
\label{appendix:crafting-quiz}

Our procedure for creating quiz options involves generating four word-level perturbations for each instance subsampled from a particular dataset partition. For each instance, we prompt GPT-4 to generate four perturbations at once.
We repeatedly apply this process to all the subsampled instances.
The template prompt and all the rules and requirements employed to instruct GPT-4 in generating word-level perturbations are shown via an example in Figure \ref{figure:quiz-generation-prompt}. This zero-shot prompt shows the four generated word-level perturbations corresponding to the example given in Figure \ref{figure:example-of-contamination-quiz}.

\begin{figure*}[!]
\scriptsize
    \begin{minipage}{\textwidth} 
        \centering
        \begin{tikzpicture}[rounded corners=8pt, thick, text=black, text opacity=1]
            \node[draw=solid_gray, fill=light_gray, line width=1pt, text=black, text width=0.97\textwidth, align=left, font=\fontsize{8pt}{1em}\selectfont, inner xsep=6.5pt, inner ysep=5pt] at (0,0) {\textbf{Instruction:}
            Your task is to create a four-choice quiz by replacing the words in the provided ``Input Text'' with their contextually relevant synonyms.
            The meaning and sentence structure of the four options must exactly match every detail in the Input Text.
            You must not include the provided Input Text as an option. \\
            
            \vspace{0.1cm}
            
            You must make sure that: \\
            (1) You generate distinct options based on the provided Input Text; \\
            (2) The only difference between options is word-level perturbations. \\
            (3) Options are ordered; \\
            (4) There is not any extra explanation; \\
            (5) You follow the following ``Format'' to generate options; \\
            (6) You comply with every specific symbol and letter detail in the given Input Text; and \\
            (7) All options retain the exact label from the Input Text, if there is one. \\

            \vspace{0.05cm}

            -- -- -- 

            \vspace{0.05cm}
            
            \textbf{Input Text:}
            \vspace{0.05cm}
            
            Summary: Premier Inn has announced it is investing £35m in opening three new hotels and revamping two others in Edinburgh.

            \vspace{0.05cm}
            
            -- -- --

            \vspace{0.05cm}
            
            \textbf{Format:}
            \vspace{0.15cm}
            
            A) \\
            Summary: \\

            \vspace{0.2cm}

            B) \\
            Summary: \\

            \vspace{0.2cm}

            C) \\
            Summary: \\

            \vspace{0.2cm}

            D) \\
            Summary: \\

            \vspace{0.05cm}

            -- -- --

            \highlight{\texttt{A)}}

            \vspace{-0.24cm}

            \highlight{\texttt{Summary: Premier Inn has revealed it is allocating £35m for the opening of three new hotels and the}}

            \vspace{-0.24cm}
            \highlight{\texttt{refurbishment of two others in Edinburgh.}}

            \highlight{\texttt{B)}}
            
            \vspace{-0.24cm}

            \highlight{\texttt{Summary: Premier Inn has stated it is committing £35m to launch three new hotels and upgrade two}}
            
            \vspace{-0.24cm}

            \highlight{\texttt{others in Edinburgh.}}

            \highlight{\texttt{C)}}

            \vspace{-0.24cm}

            \highlight{\texttt{Summary: Premier Inn has declared it is investing £35m to set up three new hotels and modernize two}}

            \vspace{-0.24cm}
            \highlight{\texttt{others in Edinburgh.}}


            \highlight{\texttt{D)}}

            \vspace{-0.24cm}

            \highlight{\texttt{Summary: Premier Inn has announced it is spending £35m to open three new hotels and renovate two}}

            \vspace{-0.24cm}

            \highlight{\texttt{others in Edinburgh.}}

            };
        \end{tikzpicture}
    \end{minipage}
    \caption{The zero-shot prompt employed for generating four word-level perturbations per dataset instance. The input text is replaced for each dataset instance, and GPT-4 is prompted to generate the four perturbations for each dataset instance individually at once. The format provided in the input prompt is adjusted based on a specific dataset/task, with column names in the dataset being used to denote distinct components of a dataset instance. For example, ``Summary'' is a column name in the XSum that is prepended to the instance shown above. The example shown here is an instance taken from the XSum test partition along with its generated word-level perturbations, as illustrated earlier in Figure \ref{figure:example-of-contamination-quiz}.}
    \label{figure:quiz-generation-prompt}
\end{figure*}

Additionally, to evaluate the quality of the generated word-level perturbations by GPT-4, we conduct a three-point Likert scale analysis on a subsample of 60 instances. These instances are drawn from four distinct datasets/tasks: AG News \citep{Zhang2015CharacterlevelCN}, XSum \citep{Narayan2018DontGM}, WNLI \citep{wang2018glue}, and GSM8k \citep{cobbe2021GSM8k}, with each dataset contributing 15 instances.
Three domain experts participate in evaluating the generated perturbations based on two independent criteria: \emph{meaning} and \emph{sentence structure}, using the following scale: {\em good}, {\em fair}, and {\em poor}.


The criteria for this evaluation are defined as:

\begin{itemize}[itemsep=0pt, topsep=0pt]
    \item \textbf{Meaning:} This criterion ensures that perturbations are solely at the word level while preserving the similarity in core essence between the generated perturbations and the original dataset instance.
    \item \textbf{Sentence Structure:}  This criterion determines the level of correspondence in sentence structure between the generated perturbations and the original dataset instance while maintaining strict adherence to every precise symbol and letter detail featured in the original dataset instance.
\end{itemize}

\vspace{0.15cm}

Additionally, the Likert scale points are described as follows:

\begin{itemize}[itemsep=0pt, topsep=0pt]
    \item \textbf{Good:} The word-level perturbations meet all aforementioned requirements for meaning and sentence structure independently.
    \item \textbf{Fair:} The word-level perturbations fulfill some, but not all, requirements for meaning and sentence structure independently.
    \item \textbf{Poor:} The word-level perturbations do not meet any of the stated requirements for meaning and sentence structure independently.
\end{itemize}

\vspace{0.15cm}

We evaluate the inter-rater reliability among expert raters by calculating the overall percent agreement, per rating percent agreement, and the corresponding confidence intervals for each criterion under consideration (i.e., \textit{meaning} and \textit{sentence structure}). 
The percent agreement represents the proportion of instances where all raters concur (on the same rating/category) out of the total number of ratings provided.
To determine the confidence intervals, we employ bootstrap resampling \citep{10.1214/aos/1176344552,EfroTibs93, 10.1214/ss/1063994968} with 10,000 iterations to estimate the 95\% confidence intervals for percent agreement for each of the ratings (i.e., {\em good}, {\em fair}, and {\em poor}).
Note that we opted not to use Cohen's/Fleiss' Kappa \citep{Cohen1960ACO,scott1955reliability,fleiss1971measuring} as our inter-rater reliability measure due to the high chance agreement in this task, attributed to the low variability in the assigned ratings.

\pgfplotsset{compat=1.7}
\begin{figure*}[!t]
\centering
\begin{tikzpicture}
\begin{groupplot}[
    group style={
    group size=2 by 1,
    horizontal sep=1.4cm,
    vertical sep=0.7cm,
    },
    width=0.3\textwidth,  
    height=0.15\textwidth,
    ybar,
    symbolic x coords={Poor, Fair, Good},
    xtick=data,
    nodes near coords,
    tick label style={font=\tiny},
    nodes near coords style={font=\tiny},
    nodes near coords align={vertical},
    axis lines*=left,
    ymin=0,
    ymax=100,
    ytick={0,100},
    title style={at={(axis description cs:0.5,-1.45)}, align=center, font=\fontsize{7}{8.7}\selectfont},
    yticklabel={\pgfmathparse{\tick}\pgfmathprintnumber{\pgfmathresult}\%},
]

\nextgroupplot[title={{Criterion: Meaning}}]
\addplot[pattern=checkerboard, pattern color=orange] coordinates {(Poor,1.67) (Fair,3.33) (Good,95.00)};
\nextgroupplot[title={{Criterion: Sentence Structure}}]
\addplot[pattern=checkerboard, pattern color=orange] coordinates {(Poor,0) (Fair,0) (Good,100)};
\end{groupplot}
\end{tikzpicture}
\caption{Percentage distribution of evaluation results for meaning and sentence structure of the generated word-level perturbations based on majority voting among three expert raters. The most common evaluation is {\em good} for both studied criteria.}
\label{figure:perturbation-quality}
\end{figure*}

\vspace{0.15cm}

The following presents the results for the overall percent agreement and per rating percent agreement for the \emph{meaning} criterion:

\begin{itemize}[itemsep=0pt, topsep=0pt]
    \item The overall percent agreement among all three expert raters is 86.67\%, with a 95\% confidence interval of 76.67\%--95.00\%. 
    \item The percent agreement for the {\em good} rating among all three expert raters is 83.33\%, with a 95\% confidence interval of 73.33\%--91.67\%. 
    \item The percent agreement for the {\em fair} rating among all three expert raters is 3.33\%, with a 95\% confidence interval of 0.00\%--8.33\%. 
    \item The percent agreement for the {\em poor} rating among all three expert raters is 0.00\%, with a 95\% confidence interval of 0.00\%--0.00\%. 
\end{itemize}

\vspace{0.15cm}

Also, the following outlines the results of the percent agreements for the \emph{sentence structure} criterion:

\begin{itemize}[itemsep=0pt, topsep=0pt]
    \item The overall percent agreement among all three expert raters is 93.33\%, with a 95\% confidence interval of 86.67\%--98.33\%. 
    \item The percent agreement for the {\em good} rating among all three expert raters is 93.33\%, with a 95\% confidence interval of 86.67\%--98.33\%. 
    \item The percent agreement for the {\em fair} rating among all three expert raters is 0.00\%, with a 95\% confidence interval of 0.00\%--0.00\%. 
    \item The percent agreement for the {\em poor} rating among all three expert raters is 0.00\%, with a 95\% confidence interval of 0.00\%--0.00\%.
\end{itemize}

Note that the percent agreement for the \emph{fair} and \emph{poor} ratings is substantially low for both criteria because perturbations were consistently evaluated as \emph{good} by raters, resulting in low variability in ratings. Hence, while the studied subset offers reliable agreement for the predominant rating (\emph{good}), the agreement on less common ratings (\emph{fair} and \emph{poor}) still requires a larger subset to be meaningful.

As an additional assessment, we implement a majority voting among the ratings provided by the raters. 
In this evaluation, if a tie occurs, we consider the lower quality recognized by the raters.
Figure \ref{figure:perturbation-quality} illustrates the percentage distribution for each criterion upon conducting majority voting.
The results reveal that the majority of the distribution across the three ratings (i.e., {\em good}, {\em fair}, and {\em poor}) is rated as {\em good} for both \textit{meaning} and \textit{sentence structure}, at 95.00\% and 100\%, respectively.
As a result, based on the majority vote, the most common evaluation for both criteria for the generated perturbations is {\em good}. In this context, {\em good}, as defined by the provided instruction, indicates that perturbations convey the same meaning as the original dataset instance while only exhibiting differences at the word level. Additionally, {\em good} for sentence structure implies that perturbations adhere to every exact symbol and letter aspect within the original dataset instance.

Therefore, based on the comprehensive analysis provided, we conclude that GPT-4 consistently generates word-level perturbations of satisfactory quality in terms of both \textit{meaning} and \textit{sentence structure.} These perturbations meet the criteria outlined in our methodology for establishing a setting in which memorization can be induced through quiz options effectively.

\section{Systematic Contamination of GPT-3.5 and Llama 2}
\label{appendix:continued-training-for-contamination}

We systematically contaminate the GPT-3.5 base model by submitting a fine-tuning job for three epochs through the OpenAI API, while keeping all other hyperparameters at their default values. 
For each of the two datasets we use in this experiment, the GPT-3.5 base model is contaminated separately, resulting in two distinct checkpoints, each corresponding to one dataset.
Although the model provider refers to this option as fine-tuning, our methodology deviates from the standard fine-tuning procedure. In fact, our goal here is to emulate the data exposure that LLMs potentially experienced during their pre-training phase. To this end, we employ data formats used in the study by \citet{golchin2024time} to introduce contamination. 
It is also important to emphasize that this type of training differs from instruction fine-tuning, as we do not infuse any specific instructions within the data we use for continued training.

To contaminate Llama 2, we use low-rank adaptation \citep{hu2022lora} for continued training, following the hyperparameter recommendations provided by the model developer \citep{llama_fine_tuning}. However, we adjust two hyperparameters to suit our use case: we set the number of epochs to three and limit the maximum input length to 500 tokens. We also use the same data formats mentioned above for contaminating the GPT-3.5 base model.

\begin{table*}[!t]
\centering
\renewcommand{\arraystretch}{1.3} 
\caption{Positional biases detected using the Bias Detector Quiz (BDQ) across all datasets and models in this work. Although these proportions may vary slightly across different runs due to the probabilistic nature of LLMs, the overall biases remain consistent, as discussed in Section \ref{section:empirical-results-and-evaluation} and shown in Figure \ref{figure:grid-of-bars}. The number next to each option represents the percentage of times that option was selected in BDQ.}
\label{tab:positional-biases}
\begin{adjustbox}{max width=\textwidth}
  \begin{tabular}{l c c c}
    \toprule
    {\large\bfseries Dataset}
      & {\large\bfseries GPT-3.5}
      & {\large\bfseries GPT-4}
      & {\large\bfseries Llama 2} \\
    \midrule
    MeetingBank      & A: 53, B: 3, C: 22, D: 1, E: 21 
                     & – 
                     & A: 67, B: 2, C: 31, D: 0, E: 0 \\
    AuTexTification  & A: 49, B: 5, C: 34, D: 0, E: 12 
                     & – 
                     & A: 41, B: 2, C: 57, D: 0, E: 0 \\
    HumanEval        & – 
                     & A: 50.61, B: 3.66, C: 0, D: 2.44, E: 43.29 
                     & – \\
    DROP             & – 
                     & A: 36, B: 5, C: 3, D: 2, E: 54 
                     & – \\
    GSM8k            & – 
                     & A: 44, B: 2, C: 1, D: 1, E: 52 
                     & – \\
    IMDB             & A: 55, B: 0, C: 16, D: 0, E: 29 
                     & A: 68, B: 1, C: 0, D: 0, E: 31 
                     & A: 33, B: 10, C: 57, D: 0, E: 0 \\
    AG News (train)  & – 
                     & A: 29, B: 0, C: 0, D: 0, E: 71 
                     & – \\
    AG News (test)   & A: 19, B: 8, C: 36, D: 3, E: 34 
                     & A: 41, B: 0, C: 1, D: 1, E: 57 
                     & A: 19, B: 7, C: 74, D: 0, E: 0 \\
    Yelp             & A: 39, B: 0, C: 22, D: 1, E: 38 
                     & A: 28, B: 0, C: 0, D: 0, E: 72 
                     & A: 40, B: 1, C: 59, D: 0, E: 0 \\
    RTE              & A: 0,  B: 0, C: 17, D: 2, E: 81 
                     & A: 16, B: 0, C: 0, D: 1, E: 83 
                     & A: 56, B: 8, C: 14, D: 0, E: 22 \\
    WNLI (train)     & – 
                     & A: 6, B: 0, C: 0, D: 0, E: 94 
                     & – \\
    WNLI (validation)& A: 0,   B: 0,   C: 5.63, D: 0,    E: 94.37 
                     & A: 9.86, B: 0,   C: 0,    D: 1.41, E: 88.73 
                     & A: 12.68, B: 26.76, C: 0,    D: 0,    E: 60.56 \\
    SAMSum           & A: 3,  B: 7,  C: 49, D: 13, E: 28 
                     & A: 15, B: 3,  C: 1,  D: 4,  E: 77 
                     & A: 63, B: 2,  C: 34, D: 0,  E: 1 \\
    XSum             & A: 16, B: 3,  C: 28, D: 5,  E: 48 
                     & A: 41, B: 7,  C: 2,  D: 8,  E: 42 
                     & A: 71, B: 0,  C: 29, D: 0,  E: 0 \\
    \bottomrule
  \end{tabular}
\end{adjustbox}
\end{table*}

\section{Statistics on Positional Biases}
\label{sec:positional-biases}

Table \ref{tab:positional-biases} presents the detected positional biases by the Bias Detector Quiz (BDQ) among all the models and datasets used in this study. The results reveal that LLMs exhibit different positional biases depending on the dataset, indicating that these biases are neither universal nor transferable across models and datasets. Moreover, this further underscores the importance of addressing positional biases when estimating contamination is necessary. For instance, although the MeetingBank and AuTexTification datasets were unseen by Llama 2, positional bias prevented the model from choosing option E---none of the options---which is the correct choice in BDQ, where no correct answer exists among options. Instead, the model disproportionately favored options A and C beyond random chance. As explained in Subsection \ref{subsection:positional-adversarial-analysis}, failing to properly handle such biases results in overestimating contamination.

\end{document}